\documentclass[conference]{IEEEtran}
\IEEEoverridecommandlockouts
\usepackage{cite}
\usepackage{amsmath,amssymb,amsfonts}
\usepackage{algorithm, algorithmic}
\usepackage{graphicx}
\usepackage{textcomp}
\usepackage[table,xcdraw]{xcolor}
\usepackage{multirow}
\usepackage{url}
\usepackage[caption=false, font=footnotesize]{subfig}

\def\BibTeX{{\rm B\kern-.05em{\sc i\kern-.025em b}\kern-.08em
    T\kern-.1667em\lower.7ex\hbox{E}\kern-.125emX}}
\begin{document}
\title{Infrastructure-Assisted Collaborative Perception in Automated Valet Parking: A Safety Perspective}

\author{\IEEEauthorblockN{Yukuan Jia\IEEEauthorrefmark{1}, 
Jiawen Zhang\IEEEauthorrefmark{1}, 
Shimeng Lu\IEEEauthorrefmark{2}, 
Baokang Fan\IEEEauthorrefmark{1},
Ruiqing Mao\IEEEauthorrefmark{1}, 
Sheng Zhou\IEEEauthorrefmark{1}, 
Zhisheng Niu\IEEEauthorrefmark{1}}

\IEEEauthorblockA{\IEEEauthorrefmark{1}Department of Electronic Engineering, Tsinghua University, Beijing, China\\
Email: \{jyk20, jiawen-z20, fbk22, mrq20\}@mails.tsinghua.edu.cn, \{sheng.zhou, niuzhs\}@tsinghua.edu.cn}

\IEEEauthorblockA{\IEEEauthorrefmark{2}School of Electronic Information and Communications, Huazhong University of Science and Technology, Wuhan, China
\\Email: u202113939@hust.edu.cn}

\thanks{This work is sponsored in part by Nature Science Foundation of China (No. 62341108, No. 62221001, No. 62022049, No. 62111530197), and by the project of Tsinghua University-Toyota Joint Research Center for AI Technology of Automated Vehicle (No. TTAD2023-08). (\textit{Yukuan Jia and Jiawen Zhang contributed equally to this work.})}
}

\maketitle

\begin{abstract}
Environmental perception in Automated Valet Parking (AVP) has been a challenging task due to severe occlusions in parking garages.
Although Collaborative Perception (CP) can be applied to broaden the field of view of connected vehicles, the limited bandwidth of vehicular communications restricts its application.
In this work, we propose a BEV feature-based CP network architecture for infrastructure-assisted AVP systems. 
The model takes the roadside camera and LiDAR as optional inputs and adaptively fuses them with onboard sensors in a unified BEV representation.
Autoencoder and downsampling are applied for channel-wise and spatial-wise dimension reduction, while sparsification and quantization further compress the feature map with little loss in data precision.
Combining these techniques, the size of a BEV feature map is effectively compressed to fit in the feasible data rate of the NR-V2X network.
With the synthetic AVP dataset, we observe that CP can effectively increase perception performance, especially for pedestrians. 
Moreover, the advantage of infrastructure-assisted CP is demonstrated in two typical safety-critical scenarios in the AVP setting, increasing the maximum safe cruising speed by up to 3m/s in both scenarios.
\end{abstract}

\section{Introduction}
% Automated Valet Parking
It is reported that tens of thousands of crashes occur in parking facilities annually in the US, causing hundreds of deaths every year \cite{nsc}.
To park a car in a garage, one needs to carefully maneuver in a tight space with imperfect lighting conditions and severe occlusions, which poses major challenges to human drivers.
With the rapid development of automated driving technologies, Automated Valet Parking (AVP) has been proposed and demonstrated with prototypes \cite{avp-survey, parkmycar}.
When AVP is enabled at the drop-off spot, the vehicle will autonomously drive through the parking lot and park itself in an empty slot.
Subsequently, the vehicle drives itself to the specified pick-up spot upon the calling of the driver.

\begin{figure}[!t]
    \centering
    \subfloat[]{\includegraphics[width=0.27\textwidth]{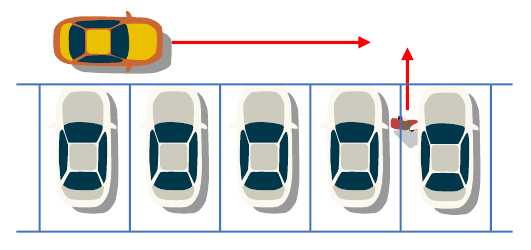}
    \label{fig:scenario1}}
    \hfill
    \subfloat[]{\includegraphics[width=0.27\textwidth]{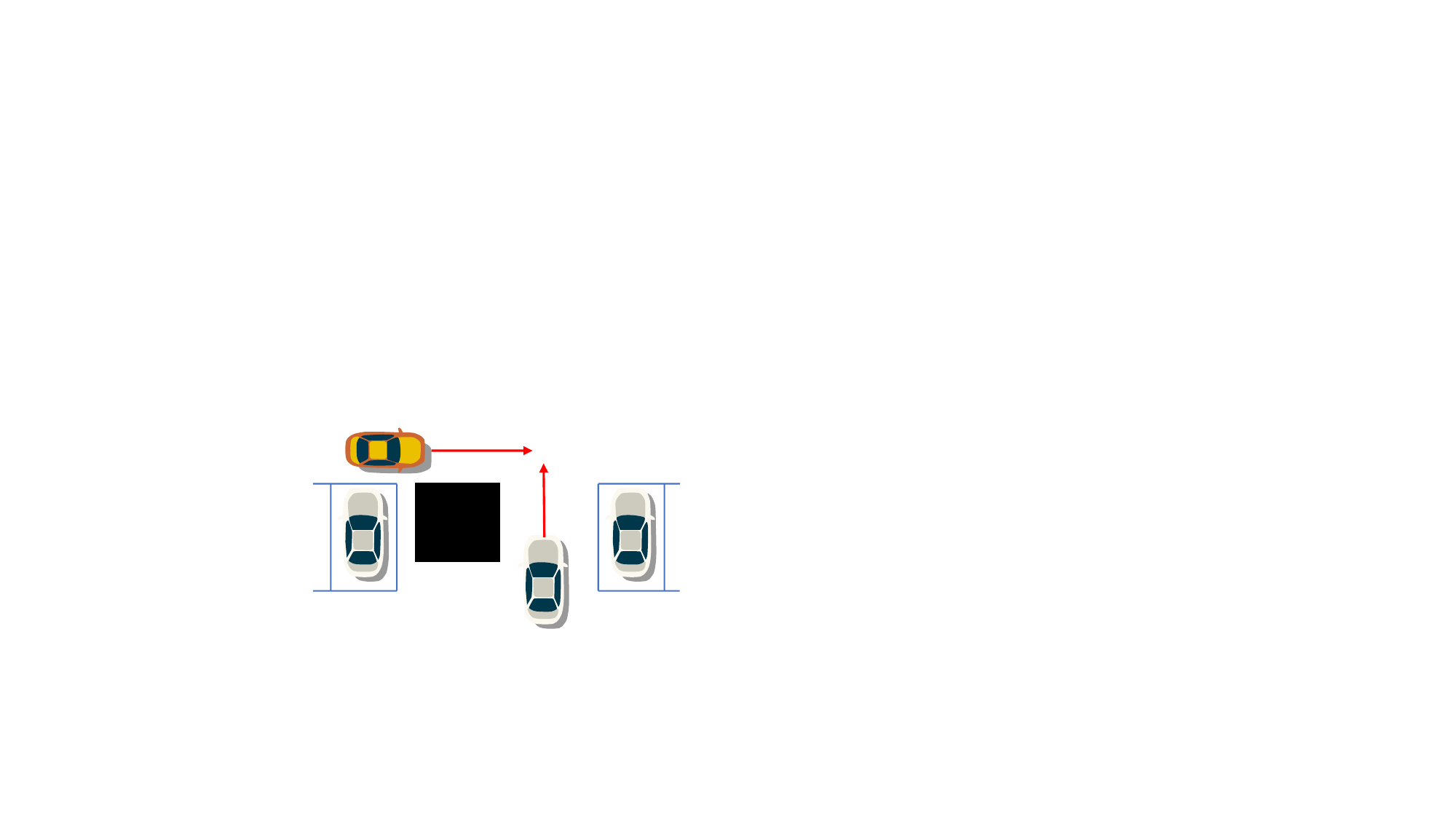}
    \label{fig:scenario2}}
    \caption{Two typical safety-critical scenarios in parking garages. (a) Crossing Pedestrian. (b) Occluded T-junction.}
    \label{fig:scenarios}
\end{figure}

% using Onboard Sensors, Critical Safety Scenarios, restrict the cruising speed, efficiency
While navigating in the garage, the vehicle needs to perceive the surrounding environment using its onboard sensors, e.g. cameras and LiDARs, to avoid collisions.
However, the perception of onboard sensors is inevitably restricted by occlusions, limited to their line-of-sight (LoS).
In this paper, two typical safety-critical scenarios caused by occlusions are identified, which could potentially lead to accidents.
As illustrated in Fig. \ref{fig:scenario1}, the yellow vehicle cannot detect the pedestrian walking out behind a parked vehicle, while in Fig. \ref{fig:scenario2}, the two vehicles are unaware of each other due to the occlusion caused by the pillar in black.
To guarantee safety in those corner cases, the vehicle must drive cautiously at a very low speed, which in turn negatively affects the operating efficiency of the parking garages.

% Connected Intelligent Vehicles, IA Cooperative Perception
Inspired by collaborative perception (CP) widely investigated in open roads \cite{cooper, tits-multi-schemes}, we propose an infrastructure-assisted CP solution to address the safety issues of AVP.
With sensors equipped in critical areas, the parking infrastructure can transmit perceptual data to vehicles via vehicular communications.
In this manner, the blind zones in the field of view of vehicles are eliminated, and the perception reliability of connected vehicles is essentially improved.
While the cellular network may not realize its full potential in a roofed garage, the wireless network can be supported by the NR-V2X protocol \cite{nr-v2x} that enables direct infrastructure-to-vehicle (I2V) communications.
In the absence of GNSS signal in an underground or multi-story garage, the synchronization reference of NR-V2X can be provided by the infrastructure~\cite{v2x-sync}.

% Challenges: lowBW, delay, inacurate gps 
There are unique challenges of infrastructure-assisted CP inside a parking garage.
Firstly, since a vehicle cannot receive GNSS signals for accurate self-localization, the position offsets could be more than 1 meter in a parking structure \cite{avp-survey}.
On the other hand, feature matching-based localization can achieve sub-meter accuracy but frequently fails at low illumination levels \cite{local-match}.
Therefore, when merging object-level information (i.e. bounding boxes) from the infrastructure, the vehicle struggles to determine their precise relative coordinates.
Secondly, the allocated spectrum of V2X communication is 20MHz in China, and 30MHz in the US \cite{v2x-spec}.
Therefore, the bandwidth can hardly accommodate raw sensor data such as HD images or LiDAR point clouds, which take tens of megabytes per second at a sampling rate of 10Hz.
This huge amount of data would also cause an increase in the end-to-end delay, which impairs the reliability of CP.

% Solutions: feature-level CP, with feature compression & registration
To solve the above challenges, we propose to transmit the intermediate bird-eye-view (BEV) feature map \cite{BEVfusion}, which is a unified representation of different sensor perspectives and modalities.
The BEV feature preserves the key foreground information while effectively reduces the transmission load.
The BEV feature map also contains spatial semantics that can be used to register and correct the coordinate transformation when fusing data from multiple sources \cite{CoAlign}. 
Moreover, the BEV maps can be compressed in many ways, such as channel-wise reduction, spatial-wise down-sampling, and leveraging sparsity.

% Contributions
In this paper, we focus on the autonomous driving part of AVP from a safety perspective. 
The infrastructure-assisted CP with different settings is tested in two typical safety-critical scenarios.
To the best of our knowledge, this is the first paper investigating infrastructure-assisted CP systems in the scenario of AVP.
There is another work \cite{raw-avp} studying a raw-level CP system where LiDAR point clouds are exchanged among vehicles but does not consider the limited wireless bandwidth. 
Our main contributions are summarized as follows:

1) A BEV feature-based CP architecture is proposed in the framework of an infrastructure-assisted AVP system, which takes multiple modalities from multiple sources as optional inputs, and adaptively fuses them in a unified BEV representation.

2) The transmitted BEV features are effectively compressed to fit in the feasible data rate of NR-V2X communication. Autoencoder and downsampling are applied for channel-wise and spatial-wise dimension reduction, while sparsification and quantization further compress the feature map with little loss in data precision.

3) The safety performance, characterized by the maximum safe cruising speed, is tested in two typical safety-critical scenarios in a simulated environment, which verifies the effectiveness of infrastructure-based CP.

The rest of this paper proceeds as follows.
An infrastructure-assisted AVP system framework is introduced in Sect. \ref{sect:framework}. 
Then, we propose the BEV feature-based CP network architecture in Sect. \ref{sect:CPArch}.
In Sect. \ref{sect:simulation}, the model benchmark and experiments in safety-critical scenarios are presented.
Finally, the paper is concluded in Sect. \ref{sect:conclusion}.

\section{Infrastructure-Assisted AVP System Framework} \label{sect:framework}

\begin{figure}[!t]
	\centering
	\includegraphics[width=0.47\textwidth]{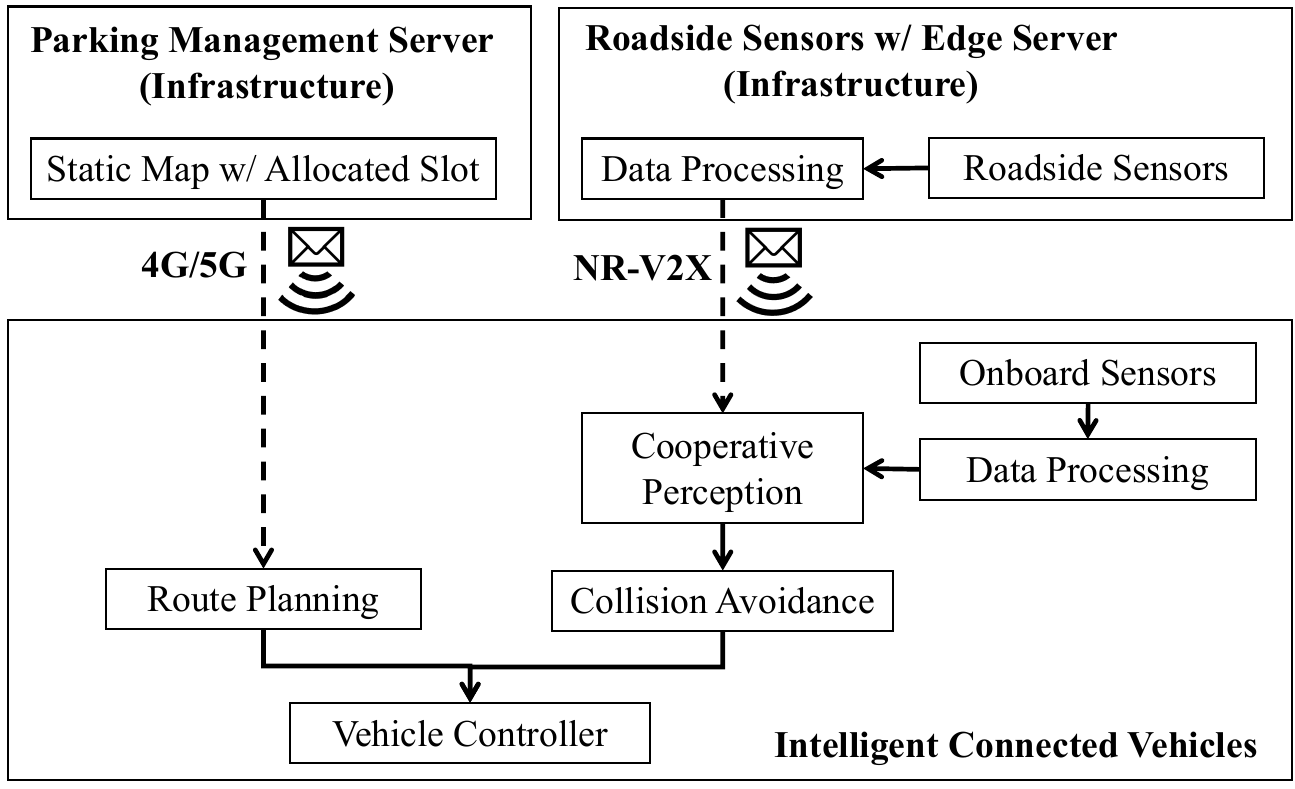}  
	\caption{The infrastructure-assisted AVP system framework.}
 	\label{fig:framework}
\end{figure}

In general, an AVP system is supported by two key modules, route planning, and collision avoidance, as illustrated in Fig. \ref{fig:framework}.
Conventionally, the vehicle needs to independently search for an empty space to park the car, which may take a long time and cause congestion.
For a smart garage, a parking management server can allocate an empty slot and meanwhile transmit the static garage map to the vehicle.
This information only needs to be transmitted once at the entrance of the garage using high-speed 4G/5G networks.
With the downloaded map, the vehicle can then determine a path to the allocated parking slot. 
There is a line of research on the optimal allocation of parking slots~\cite{opt-allocate}, but it is beyond the scope of this paper.

For collision avoidance, automated vehicles are equipped with a set of onboard sensors to perceive the surrounding environment.
A basic, low-cost onboard sensor set-up is camera-only.
Although the camera can provide high-resolution contextual information, the depth cannot be accurately recovered, which is crucial for the 3D object detection task.
On the other hand, LiDAR is much more accurate in 3D position measurements, but the point clouds are relatively sparse compared to images.
To exploit the benefits of both modalities, a more reliable but expensive solution is to fuse the images and point clouds \cite{BEVfusion, CoBEVT}.

To address the inherent limitation of onboard sensors caused by occlusions, roadside sensors can be deployed as parking infrastructure.
Similarly, the infrastructure also has two alternative set-ups, camera-only and a combination of camera and LiDAR.
The sensor data is captured and processed at the edge server and transmitted to the connected vehicles for CP via NR-V2X communication.
With a bandwidth of 20MHz in the sub-6 GHz band, the theoretical data rate of NR-V2X ranges from 19.1Mbps to 76.5Mbps, depending on the exact modulation and coding scheme \cite{v2x-rate}.
Considering the co-existence of other intelligent transportation applications, the available bandwidth is constrained to a few megabytes (MB) per second.
To fit in the limited data rate, the amount of the transmitted sensor data needs to be notably compressed, as will be further introduced in Sect. \ref{sect:compress}.

\begin{figure*}
	\centering
	\includegraphics[width=0.82\textwidth]{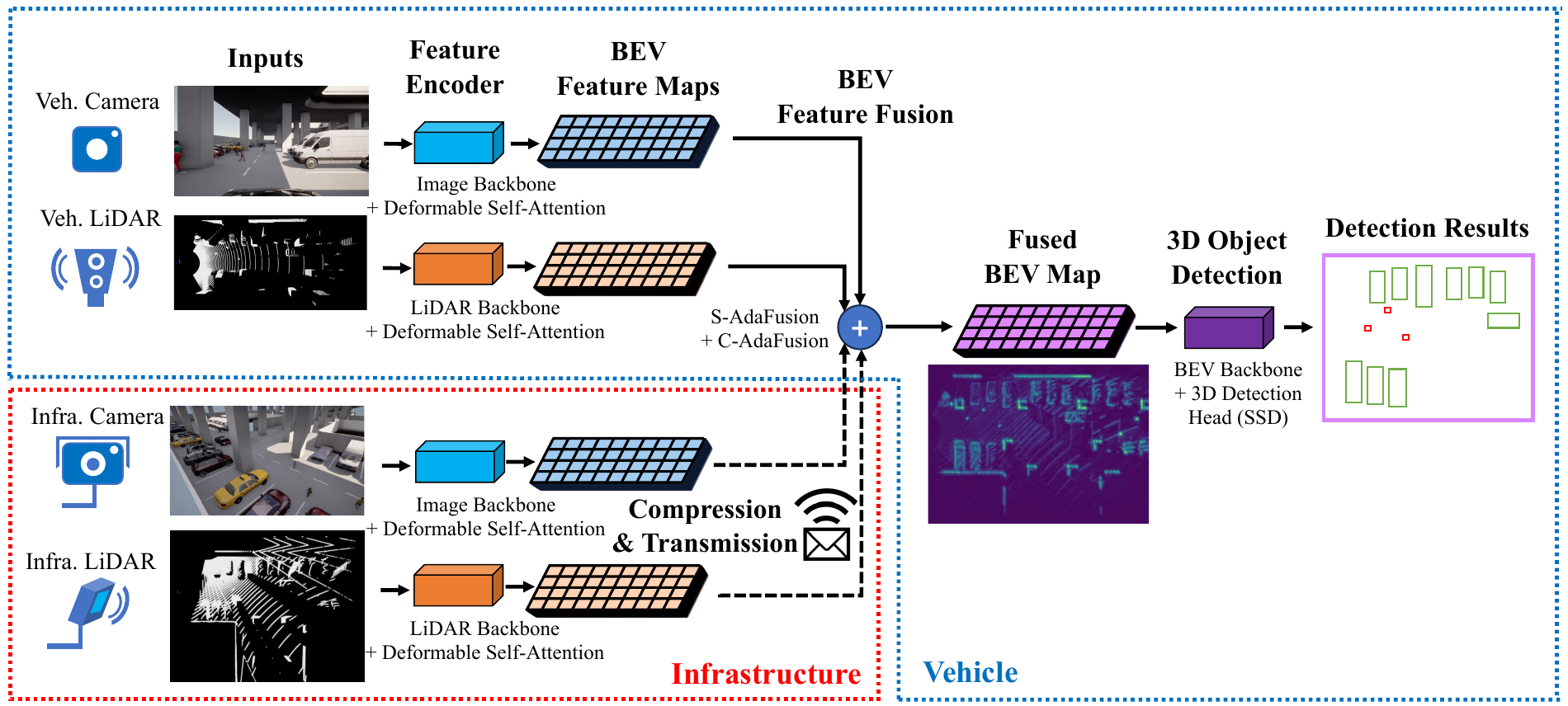}  
	\caption{The proposed BEV feature-based CP network architecture.}
 	\label{fig:CPArch}
\end{figure*}

\section{BEV Feature-based CP Network Architecture} \label{sect:CPArch}

In this section, we propose a CP network architecture shown in Fig. \ref{fig:CPArch}.
The infrastructure compresses and transmits intermediate BEV feature maps to connected vehicles.
Note that this architecture degrades gracefully to onboard-only perceptions when the infrastructure is unavailable.

\subsection{Multi-Modal Feature Encoder}
To fuse sensor data of different modalities and from different perspectives, the raw sensor data needs to be encoded to BEV features in a unified representation space.
For LiDAR point clouds, the pillar feature network in PointPillars~\cite{PointPillars} is adopted.
The 3D space is first evenly discretized into grids in the 2D BEV plane, generating the tall pillars that contain the scanned points.
Then a neural network extracts a set of features from the points for each pillar, which forms a 2D BEV feature map. 

For camera images, we introduce the Categorical Depth Distribution Network (CaDDN) \cite{CaDDN} to achieve camera-to-BEV transformation.
In the backbone, a categorical depth distribution is predicted for each image pixel, projecting the rich semantic information into the 3D frustum feature space.
With the camera calibration information, the features in the 3D frustum space are then resampled in the 3D voxel space.
Finally, the voxel features are collapsed to the 2D-grid BEV plane using a $1\times 1$ convolution network.

Since both backbones encode features independently inside each BEV grid, a deformable self-attention module \cite{DefDETR} is added to the end of each backbone.
As a complement, this module learns to encode the spatial relationship between each grid and a small set of sampling points around it.
Additionally, this also creates a unified representation space for any input modality.
% Note that these learnable feature encoders with parameters are trained in an end-to-end fashion.

\subsection{Feature Compression} \label{sect:compress}
Formally, an encoded BEV feature map $X\in\mathbb{R}^{H \times W \times C}$ has spatial dimension $H \times W$, with $C$ feature channels.
Without any compression, the default data size could be tens of megabytes for a single map, which does not match the limited bandwidth of V2X communication.
In this work, four techniques are introduced to compress the data size, in the dimensions of channel, space, data sparsity, and precision, respectively.

Similar to \cite{disconet}, we first apply an auto-encoder, i.e. a 1$\times$1 convolution network, to reduce the number of channels from $C$ to $C/8$.
Then, a 2$\times$2 max pooling layer is used to downsample the size of the BEV feature map to $ (H/2) \times (W/2)$.
Furthermore, the BEV map usually has a high sparsity \cite{PointPillars}, i.e. the majority of the grids have negligible values.
The feature map can be further compressed using a sparse tensor structure.
Finally, the classic quantization method can also be applied by using a lower-precision floating-point data type.
Combining these four compression techniques, the data amount of each BEV map can be decreased to the magnitude of kilobytes~(KB), without noticeable degradation in perception performance.

\subsection{Feature Fusion}
When the compressed BEV feature maps are received from the infrastructure, the vehicle first performs upsampling and feeds them into an auto-decoder to recover the original BEV maps $X_\text{inf}^\text{cam}$, $X_\text{inf}^\text{lidar}$.
A spatial transformation $\Gamma(\cdot)$ is then needed to project the received maps onto the vehicle's BEV coordinate system, expressed by $\Gamma\left(X_\text{inf}^\text{cam}\right)$, $\Gamma\left(X_\text{inf}^\text{lidar}\right) \in \mathbb{R}^{H \times W \times C}$.
There exist a variety of ways to fuse the BEV features maps.
% of different modalities and from different sources.
For example, the simplest method is to apply average pooling over the feature maps, given by
\begin{align}
    X_\text{avg} = \frac{1}{4}\left[X_\text{veh}^\text{cam} + X_\text{veh}^\text{lidar} + \Gamma\left(X_\text{inf}^\text{cam}\right) + \Gamma\left(X_\text{inf}^\text{lidar}\right)\right].
\end{align}
However, some unique and important features fade out when there are multiple inputs, which inspires us to fuse the BEV features with adaptive weights.

In reference \cite{AdaFusion}, a spatial-wise adaptive feature fusion (S-AdaFusion) is proposed to adaptively fuse the average pooling and the max pooling of all feature maps, i.e.,
\begin{align}
    X_\text{S-Ada} = \text{3DConv}\left(\left[X_\text{avg}, X_\text{max}\right]\right),
\end{align}
where $X_\text{max}$ is the element-wise maximization over feature maps, and the 3D convolution is utilized for feature selection and dimension reduction to one single map.
On the other hand, channel-wise adaptive feature fusion (C-AdaFusion) calculates the weight of each map by a function of two scalar map descriptors, given by
\begin{align}
    W_i = g(\text{max}(X_i), \text{avg}(X_i)).
\end{align}
Then, the weighted average is taken over all feature maps, i.e., $X_\text{C-Ada} = \sum_{i} W_i X_i$.
In our CP architecture, we propose to combine the above S-AdaFusion and C-AdaFusion modules using another C-AdaFusion module, to take advantage of both spatial-wise and channel-wise adaptivity. 
% Note that our proposed fusion method 

\subsection{3D Object Detection}
The last step is to conduct the 3D object detection on the fused BEV map.
The BEV backbone and the detection head of PointPillars \cite{PointPillars} are reused, which is well-established in the literature and has low computation complexity.

Specifically, the BEV backbone has a series of top-down convolutional networks that gradually downsample the BEV map and generate multi-scale BEV features.
Then the features are upsampled and concatenated to form the final BEV features.
The classic single shot detector (SSD) is applied to predict the positions and sizes of the bounding boxes, and the classification scores for each category.

\section{Experiments} \label{sect:simulation}
In this section, we first set up a simulated parking garage environment, collect a dataset to train the proposed CP network, and then conduct joint perception-control simulations in the safety-critical scenarios.

\subsection{Simulation Environment} 
\begin{figure}[t]
	\centering
	\includegraphics[width=0.42\textwidth]{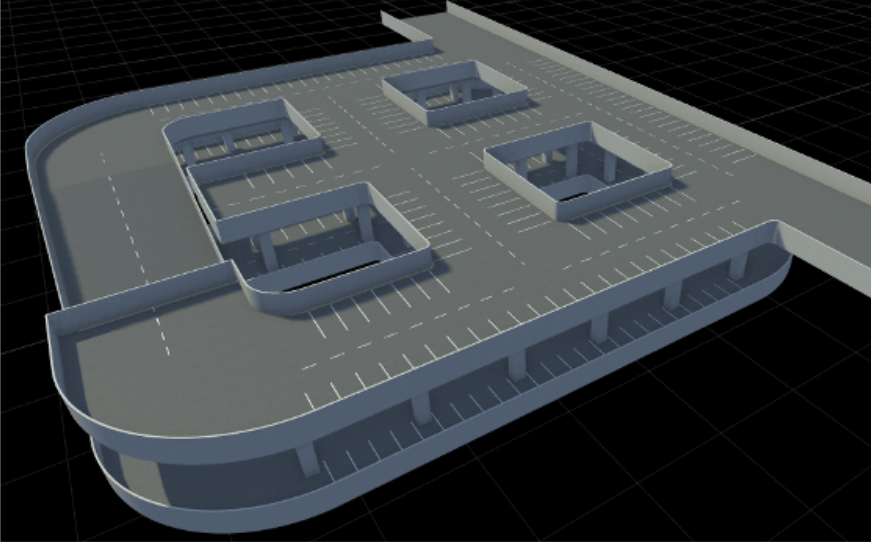}  
	\caption{Parking garage designed using Mathworks RoadRunner.}
 	\label{fig:garage}
\end{figure}

We built a two-floor virtual parking garage using Mathworks RoadRunner, as shown in Fig. \ref{fig:garage}.
In this garage, the size of a parking slot is 5.8m$\times$3.2m, which can accommodate a car or a regular-sized van. 
The internal road is dual-lane with uncoordinated intersections, and the lane length is 3.4 meters. 
We focus on one straight road, including the T-junction, at Floor B1 of the garage, because the support pillars create additional occlusions which is challenging for the perception of automated vehicles.
The garage is exported as an additional map to the open-source CARLA simulator \cite{carla}, which features the rendering of sensor measurements.
This enables us to conduct a joint perception-control simulation to test the safety of AVP systems.

The AVP vehicle is equipped with a front RGB camera, and an optional 360\textdegree ~surrounding LiDAR on the top, which forms two possible setups for vehicular perception.
There are three alternative configurations for roadside infrastructure sensors, none, camera-only, and the camera-LiDAR combination.
The infrastructure sensors are set at 7.0 meters height near the T-junction, facing downwards at an angle of 30\textdegree.
% Examples of the camera images at vehicle and infrastructure are shown in Fig. \ref{fig:example-image}.
The parameter settings of the sensors are listed in Table \ref{table:param}.

% \begin{figure}[!t]
%     \centering
%     \subfloat[]{\includegraphics[width=0.4\textwidth]{figs/veh_cam.pdf}}
%     \label{fig:veh_cam}
%     \hfill
%     \subfloat[]{\includegraphics[width=0.4\textwidth]{figs/infra_cam.pdf}}
%     \label{fig:infra_cam}
%     \caption{Examples of captured images from (a) the onboard camera, and (b) the roadside camera.}
%     \label{fig:example-image}
% \end{figure}

\begin{table}[t]
\centering
\caption{Parameter Settings of Sensors}
\label{table:param}
\renewcommand{\arraystretch}{1.2}
\begin{tabular}{llll}
\hline
Sensor                  & Parameter                            & Vehicle       & Infrastructure \\ \hline
\multirow{4}{*}{Camera} & Image Resolution                      & 960$\times$540       & 960$\times$540        \\
                        & Horizontal Field-of-View    & 90\textdegree            & 90\textdegree             \\
                        & Vertical FoV      & 90\textdegree            & 90\textdegree             \\
                        & Height                   & 1.8m           & 7.0m            \\ \hline
\multirow{5}{*}{LiDAR}  & Number of Lasers                      & 64            & 64             \\
                        & Horizontal FoV  & 360\textdegree      & 120\textdegree            \\
                        & Vertical FoV   & -24.8\textdegree$\sim$+2\textdegree & -60\textdegree$\sim$+30\textdegree \\
                        & Angular Resolution       & 0.06\textdegree          & 0.02\textdegree           \\
                        & Dropoff Rate                          & 0.1           & 0.1            \\ \hline
\end{tabular}
\end{table}

\subsection{Model Training \& Benchmark}
Similar to \cite{dolphins}, we first collect a simulated AVP dataset, consisting of 2240 simultaneously captured camera images and LiDAR point clouds rendered from both perspectives. 
The corresponding ground-truth labels and sensor calibration information are exported from the CARLA simulator.
In the simulation, some vehicles are spawned to drive along the internal road, and others are parked at slots.
Pedestrians are randomly spawned between the parked vehicles, walking back and forth to cross the road.
All the road users within [0.0m, 44.8m] in the driving direction and [-30.08m, 30.08m] in the perpendicular direction, regardless of occlusions, are required to be detected by the AVP vehicle.
The difficulties of ground-truth labels are classified as easy, hard, and invisible, depending on the number of visible vertices of the bounding box from the perspective of the vehicle.

The dataset is randomly split into training and test sets with an 80\%:20\% ratio.
On the open-source platform OpenPCDet~\cite{openpcdet}, we implement the proposed CP network architecture with BEV grid size 0.16m$\times$0.16m and train the network for 150 epochs using the one-cycle Adam optimizer.
Since there are 2$\times$3 combinations of input sensor modalities, a random combination of sensors is chosen for each training iteration.

\begin{table*}[htbp]
\centering
\caption{Model Benchmark}
\label{table:benchmark}
\renewcommand{\arraystretch}{1.2}
\begin{tabular}{|cc|lll|lll|}
\hline
\multicolumn{2}{|c|}{Average Precision (AP) (\%)}                                  & \multicolumn{3}{c|}{Car IoU 0.7/0.01}                                                             & \multicolumn{3}{c|}{Pedestrian IoU 0.3/0.01}                                                       \\ \hline
\multicolumn{1}{|c|}{Vehicle Sensors}            & Infra. Sensors                  & \multicolumn{1}{l|}{Easy}      & \multicolumn{1}{l|}{Hard}      & Overall (w/ invisible)          & \multicolumn{1}{l|}{Easy}      & \multicolumn{1}{l|}{Hard}      & Overall (w/ invisible)           \\ \hline
\multicolumn{1}{|c|}{\multirow{3}{*}{Cam.}}       & None                            & \multicolumn{1}{l|}{85.2/98.3} & \multicolumn{1}{l|}{79.8/97.8} & 76.0/96.9                       & \multicolumn{1}{l|}{73.8/78.5} & \multicolumn{1}{l|}{68.0/73.1} & 57.2/61.8                        \\ \cline{2-8} 
\multicolumn{1}{|c|}{}                           & Cam.                            & \multicolumn{1}{l|}{98.0/99.7} & \multicolumn{1}{l|}{97.4/99.6} & 97.1/99.5 (+21.1/+2.6)          & \multicolumn{1}{l|}{81.9/84.6} & \multicolumn{1}{l|}{77.0/80.6} & 70.4/75.2 (+13.2/+13.4)          \\ \cline{2-8} 
\multicolumn{1}{|c|}{}                           & Cam.+LiDAR                      & \multicolumn{1}{l|}{98.5/99.7} & \multicolumn{1}{l|}{98.2/99.6} & \textbf{98.1/99.6 (+22.1/+2.7)} & \multicolumn{1}{l|}{93.0/93.6} & \multicolumn{1}{l|}{90.6/91.9} & \textbf{91.4/92.6 (+34.2/+30.8)} \\ \hline
\multicolumn{1}{|c|}{\multirow{3}{*}{Cam.+LiDAR}} & None                            & \multicolumn{1}{l|}{96.5/99.7} & \multicolumn{1}{l|}{96.2/99.7} & 91.6/97.2                       & \multicolumn{1}{l|}{99.4/99.7} & \multicolumn{1}{l|}{99.0/99.3} & 83.6/84.3                        \\ \cline{2-8} 
\multicolumn{1}{|c|}{}                           & Cam.                            & \multicolumn{1}{l|}{98.8/99.8} & \multicolumn{1}{l|}{98.8/99.8} & 98.5/99.7 (+6.9/+2.5)           & \multicolumn{1}{l|}{99.4/99.8} & \multicolumn{1}{l|}{98.6/99.0} & 89.2/91.7 (+5.6/+7.4)            \\ \cline{2-8} 
\multicolumn{1}{|c|}{}                           & \multicolumn{1}{l|}{Cam.+LiDAR} & \multicolumn{1}{l|}{98.8/99.8} & \multicolumn{1}{l|}{98.8/99.8} & \textbf{98.8/99.8 (+7.2/+2.6)}  & \multicolumn{1}{l|}{99.2/99.5} & \multicolumn{1}{l|}{99.3/99.5} & \textbf{98.9/99.3 (+15.3/+9.0)}  \\ \hline
\end{tabular}
\end{table*}

% + illustration of bev feature map?

The quantitative benchmark results with the test set are shown in Tab. \ref{table:benchmark}, where the overall column denotes all difficulties including the invisible objects.
For the 3D detection task, an object is detected only if the intersection-over-union~(IoU) of the predicted bounding box and the ground-truth bounding box is larger than a threshold, which is typically 0.7 for cars and 0.3 for pedestrians.
To proximate the demand for vehicular control, an IoU threshold of 0.01 is added, which relaxes the requirement for position accuracy and focuses on object identification.
From the model benchmark results, we can draw a few conclusions.
Firstly, the combination of camera and LiDAR is more reliable than the camera-only configuration, especially for pedestrians which are smaller in size and thus harder to detect.
Secondly, the infrastructure-assisted CP significantly improves the perception performance.
While the roadside camera is reliable enough for car detection, pedestrians require LiDAR to exploit the extra view to the fullest potential. 
The average precision of both cars and pedestrians reaches 99\% when LiDARs are set up both on automated vehicles and the parking infrastructure. 

Without compression, the shape of a BEV map is $376\times280$, with 64 channels in float32 data type, which takes 25.7MB per map.
By statistics, the sparsity of the BEV maps ranges from 0.82 to 0.88, which contributes a compression ratio of 6 to 9.
Moreover, the float16 data type is also readily available to substitute the float32 type.
Combining the four compression methods, the size of a single BEV map is reduced to 48KB~$\sim$~72KB. 
At a transmitting frequency of 20Hz, the required data rate is 0.96MB/s~$\sim$~1.44MB/s for one modality, which falls in the feasible range of the NR-V2X network. 

\subsection{Evaluations of Safety-Critical Scenarios}
A high AP value is not enough to guarantee safety in the actual automated driving.
We further conduct joint perception-control simulations in the two safety-critical scenarios previously shown in Fig. \ref{fig:scenarios}.

\begin{figure}[t]
	\centering
	\includegraphics[width=0.33\textwidth]{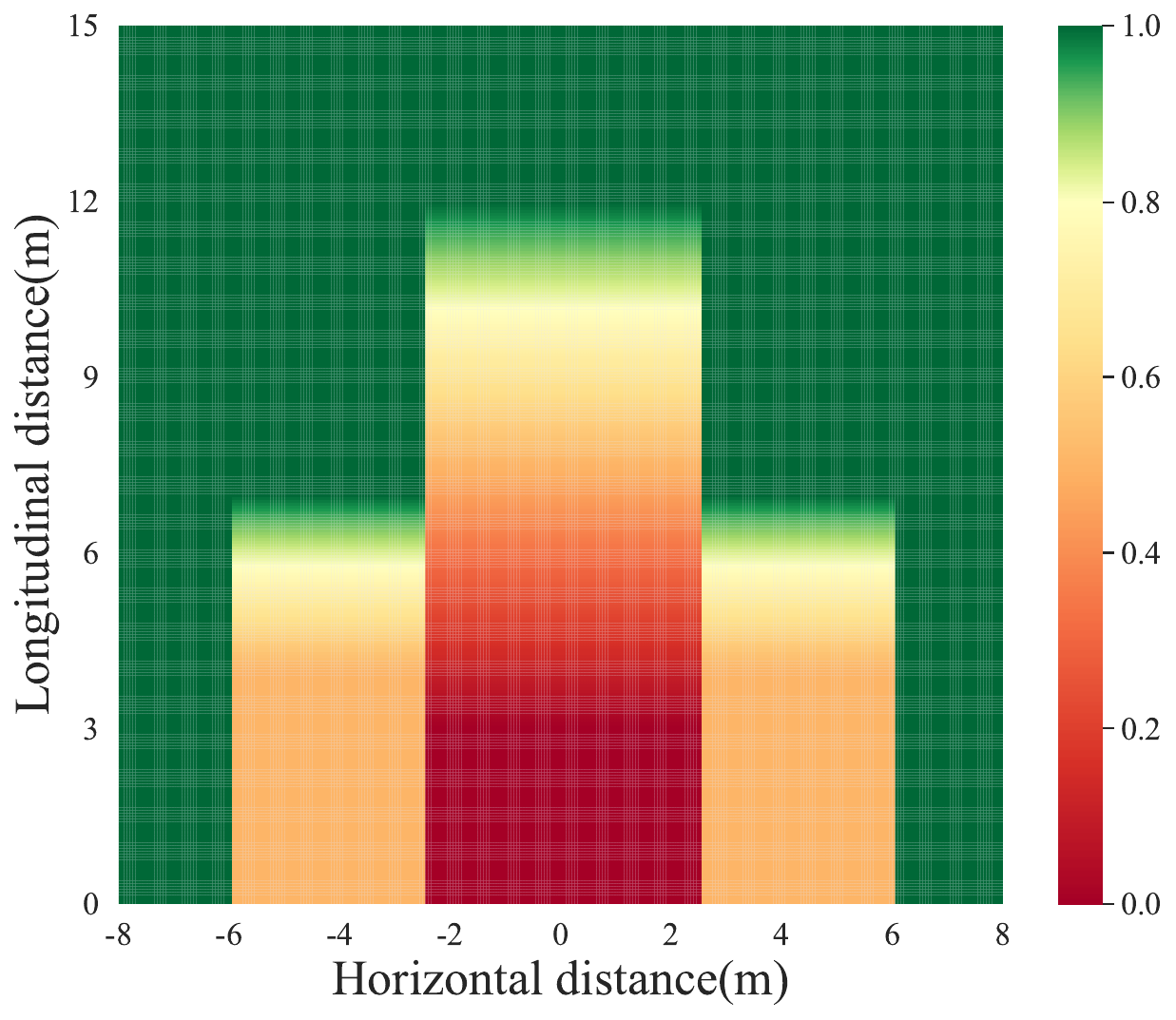}  
	\caption{Heatmap of the target speed of the AVP vehicle.}
 	\label{fig:heatmap}
\end{figure}

In the simulation, the AVP vehicle perceives the environment and determines the control decision at a frequency of 20Hz.
Given the detected walkers from the perception module, the target speed of the vehicle is set as
\begin{align}
    v^*(t) = \begin{cases}
        v_c \min\left(\max\left(\frac{\Delta x_l - 3.0}{9}, 0.0\right), 1.0\right), \Delta x_h \in [0.0,2.5], \\
        v_c \min\left(\max\left(\frac{\Delta x_l - 1.0}{6}, 0.5\right), 1.0\right), \Delta x_h \in (2.5,6.0],
    \end{cases}
\end{align}
where $v_c$ denotes the preset cruising speed inside the garage, and $\Delta x_l, \Delta x_h$ are the longitudinal and horizontal distances between the vehicle and the walker, respectively.
The heatmap of the target speed is illustrated in Fig. \ref{fig:heatmap}.
A simple linear relationship is assumed between the target acceleration and the target speed, i.e.,
\begin{align}
    a^*(t) = k v^*(t),
\end{align}
where $k=2.0 \text{s}^{-1}$ is a constant that reflects the sensitivity of vehicular control. 
An approximated three-order linear longitudinal vehicle dynamic model \cite{dynamics} is adopted for the vehicles.
The derivative of acceleration, i.e., the jerk, is given by
\begin{align}
    \dot{a}(t) = (a^*(t)-a(t))/\tau,
\end{align}
where $\tau$ is a time constant that characterizes the time lag of the powertrain system.
The actual deceleration and the jerk are restricted to no more than $\text{10.5m/s}^2$ and $\text{20m/s}^3$ for baseline automatic emergency braking (AEB) systems \cite{brake}.

\begin{table*}[htbp]
\centering
\caption{Number of Crashes in the Pedestrian Crossing Scenario}
\label{table:ped-cross}
\renewcommand{\arraystretch}{1.2}
\begin{tabular}{|cc|cccccccc|}
\hline
\multicolumn{2}{|c|}{Sensor Configurations}         & \multicolumn{8}{c|}{Cruising Speed of Vehicles}                                                                                                                                                                                                                                                                                                                               \\ \hline
\multicolumn{1}{|c|}{Vehicular}              & Infrastructure & \multicolumn{1}{c|}{3.0m/s}                                 & \multicolumn{1}{c|}{4.0m/s}                                 & \multicolumn{1}{c|}{5.0m/s}                                 & \multicolumn{1}{c|}{6.0m/s}                                 & \multicolumn{1}{c|}{7.0m/s}                                 & \multicolumn{1}{c|}{8.0m/s} & \multicolumn{1}{c|}{9.0m/s}                                 & 10.0m/s \\ \hline
\multicolumn{1}{|c|}{}                             & None           & \multicolumn{1}{c|}{{\color[HTML]{FE0000} \textbf{0/20}}} & \multicolumn{1}{c|}{1/20}                                 & \multicolumn{1}{c|}{3/20}                                 & \multicolumn{1}{c|}{3/20}                                 & \multicolumn{1}{c|}{}                                  & \multicolumn{1}{c|}{}  & \multicolumn{1}{c|}{}                                  &    \\ \cline{2-10} 
\multicolumn{1}{|c|}{}                             & Cam.           & \multicolumn{1}{c|}{}                                  & \multicolumn{1}{c|}{{0/20}} & \multicolumn{1}{c|}{{\color[HTML]{FE0000} \textbf{0/20}}} & \multicolumn{1}{c|}{1/20}                                 & \multicolumn{1}{c|}{2/20}                                 & \multicolumn{1}{c|}{2/20} & \multicolumn{1}{c|}{}                                  &    \\ \cline{2-10} 
\multicolumn{1}{|c|}{\multirow{-3}{*}{Cam.}}       & Cam.+LiDAR     & \multicolumn{1}{c|}{}                                  & \multicolumn{1}{c|}{}                                  & \multicolumn{1}{c|}{0/20}                                 & \multicolumn{1}{c|}{{\color[HTML]{FE0000} \textbf{0/20}}} & \multicolumn{1}{c|}{1/20}                                 & \multicolumn{1}{c|}{2/20} & \multicolumn{1}{c|}{1/20}                                 &    \\ \hline
\multicolumn{1}{|c|}{}                             & None           & \multicolumn{1}{c|}{}                                  & \multicolumn{1}{c|}{{}}  & \multicolumn{1}{c|}{0/20}                                 & \multicolumn{1}{c|}{{\color[HTML]{FE0000} \textbf{0/20}}} & \multicolumn{1}{c|}{1/20}                                 & \multicolumn{1}{c|}{1/20} & \multicolumn{1}{c|}{1/20}                                 &    \\ \cline{2-10} 
\multicolumn{1}{|c|}{}                             & Cam.           & \multicolumn{1}{c|}{}                                  & \multicolumn{1}{c|}{}                                  & \multicolumn{1}{c|}{{}}  & \multicolumn{1}{c|}{0/20}                                 & \multicolumn{1}{c|}{{\color[HTML]{FE0000} \textbf{0/20}}} & \multicolumn{1}{c|}{2/20} & \multicolumn{1}{c|}{1/20}                                 & 1/20  \\ \cline{2-10} 
\multicolumn{1}{|c|}{\multirow{-3}{*}{Cam.+LiDAR}} & Cam.+LiDAR     & \multicolumn{1}{c|}{}                                  & \multicolumn{1}{c|}{}                                  & \multicolumn{1}{c|}{}                                  & \multicolumn{1}{c|}{}                                  & \multicolumn{1}{c|}{{}}  & \multicolumn{1}{c|}{0/20} & \multicolumn{1}{c|}{{\color[HTML]{FE0000} \textbf{0/20}}} & 1/20  \\ \hline
\end{tabular}
\end{table*}

In the first scenario, adults and children are randomly spawn at the gaps between the parked vehicles, and cross the road without noticing the ego vehicle. 
This is to simulate the situation of an absent-minded driver or passenger leaving a parked car.
The speeds of adults and children are set to 1.5m/s and 2.0m/s respectively, and the spawn probability is 0.5 for each gap.
The vehicle, at its preset cruising speed, needs to detect the crossing pedestrian and brake in time to avoid a collision.
The various appearances and the spawn positions of parked vehicles and pedestrians create extra randomness to the scenario.

\begin{figure}[t]
    \centering
    \subfloat[]{\includegraphics[width=0.4\textwidth]{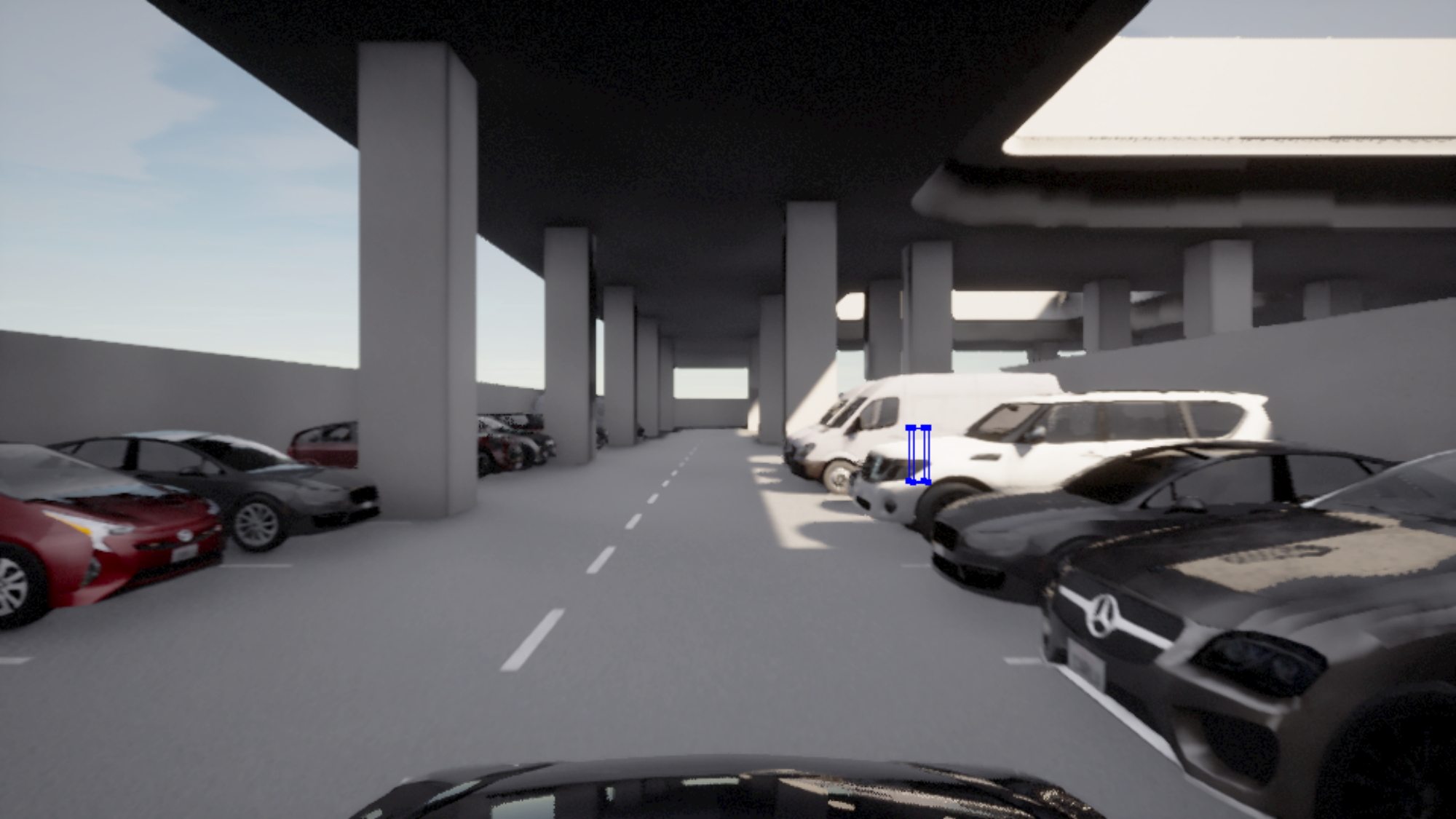}}
    \label{fig:s1_ego}
    \hfill
    \subfloat[]{\includegraphics[width=0.4\textwidth]{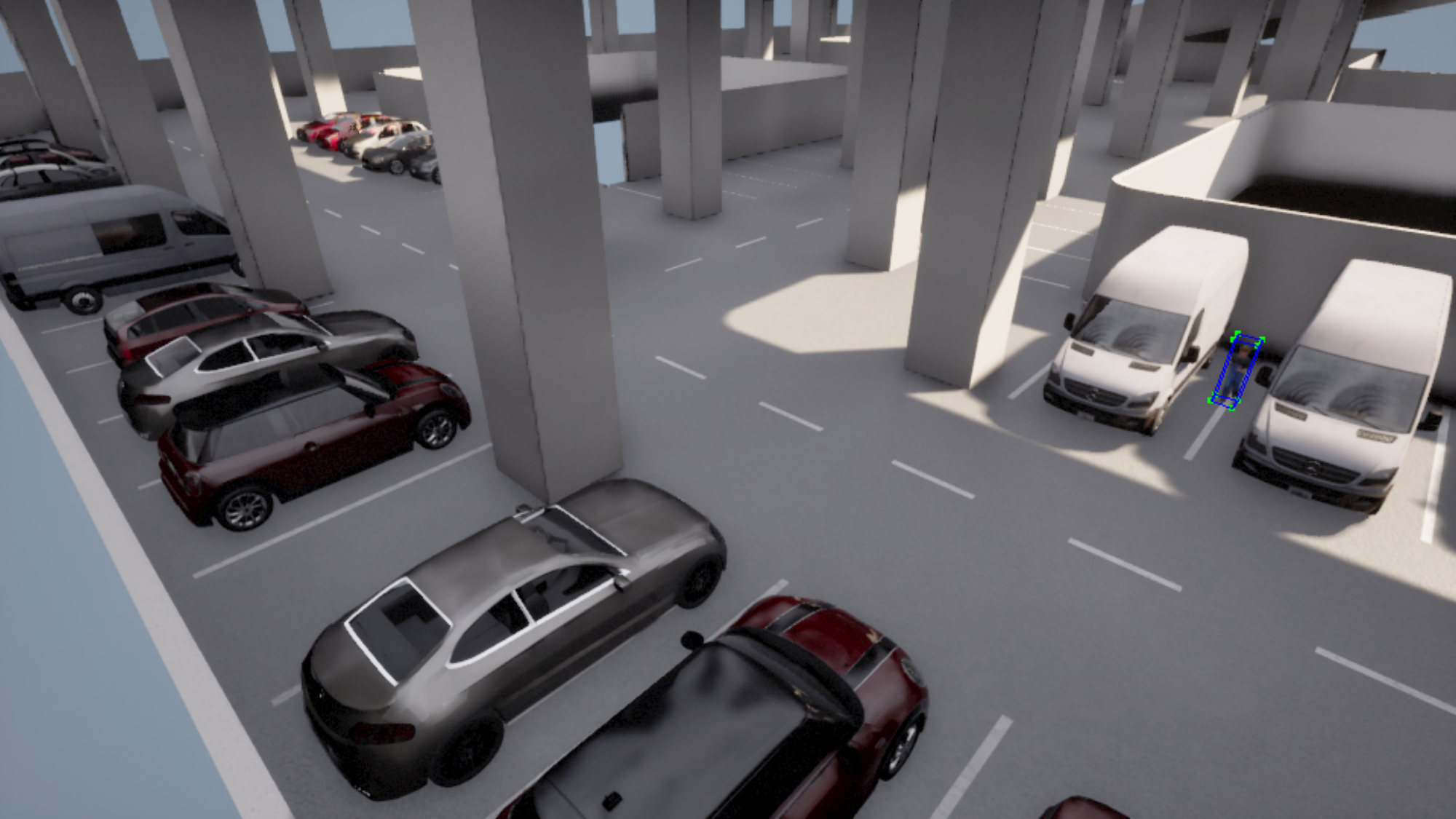}}
    \label{fig:s1_inf}
    \caption{Images from (a) the onboard camera and (b) the roadside camera in the pedestrian crossing scenario.}
    \label{fig:s1_images}
\end{figure}

The benefit of CP is qualitatively shown in Fig. \ref{fig:s1_images}, where the perspective of the roadside camera can help the AVP vehicle detect the completely occluded pedestrian between the white vans.
Tab. \ref{table:ped-cross} shows the number of crushes with different cruising speeds and sensor configurations.
The maximum safe cruising speeds for each sensor configuration are marked in red.
It can be observed that the camera-only AVP vehicle has a safe cruising speed of 3m/s, while the safe speed can be increased to 6m/s if LiDAR is available.
This is because the LiDAR is better at detecting partially occluded pedestrians than the camera.
With a roadside camera, the maximum safe speed increases to 5m/s and 7m/s, respectively.
When a combination of the roadside camera and LiDAR is applied, the maximum safe cruising speed increases by 3m/s for both vehicle configurations, which can dramatically increase the potential operating efficiency of the parking garage.

\begin{figure}[t]
    \centering
    \subfloat[]{\includegraphics[width=0.4\textwidth]{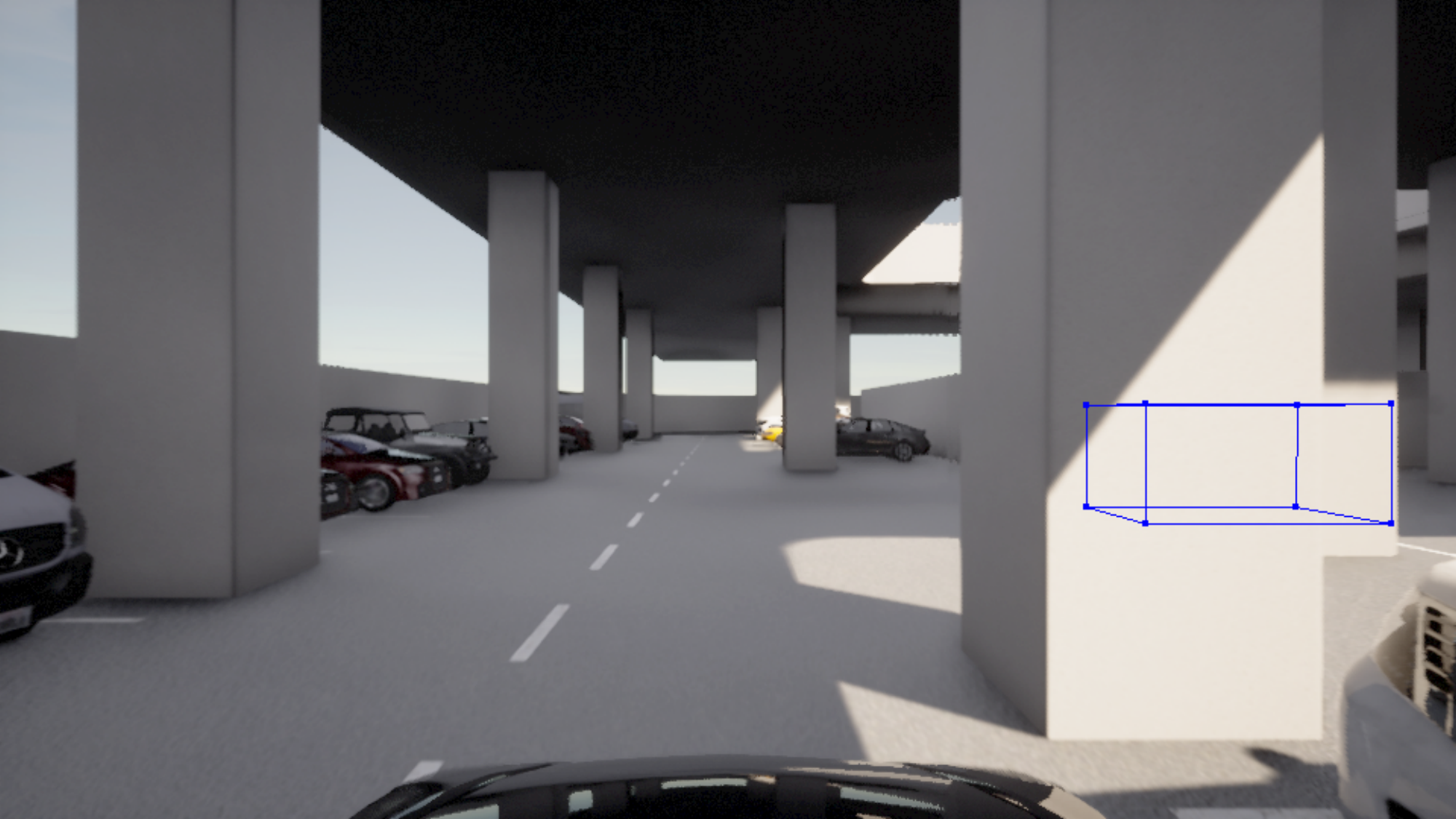}}
    \label{fig:s2_ego}
    \hfill
    \subfloat[]{\includegraphics[width=0.4\textwidth]{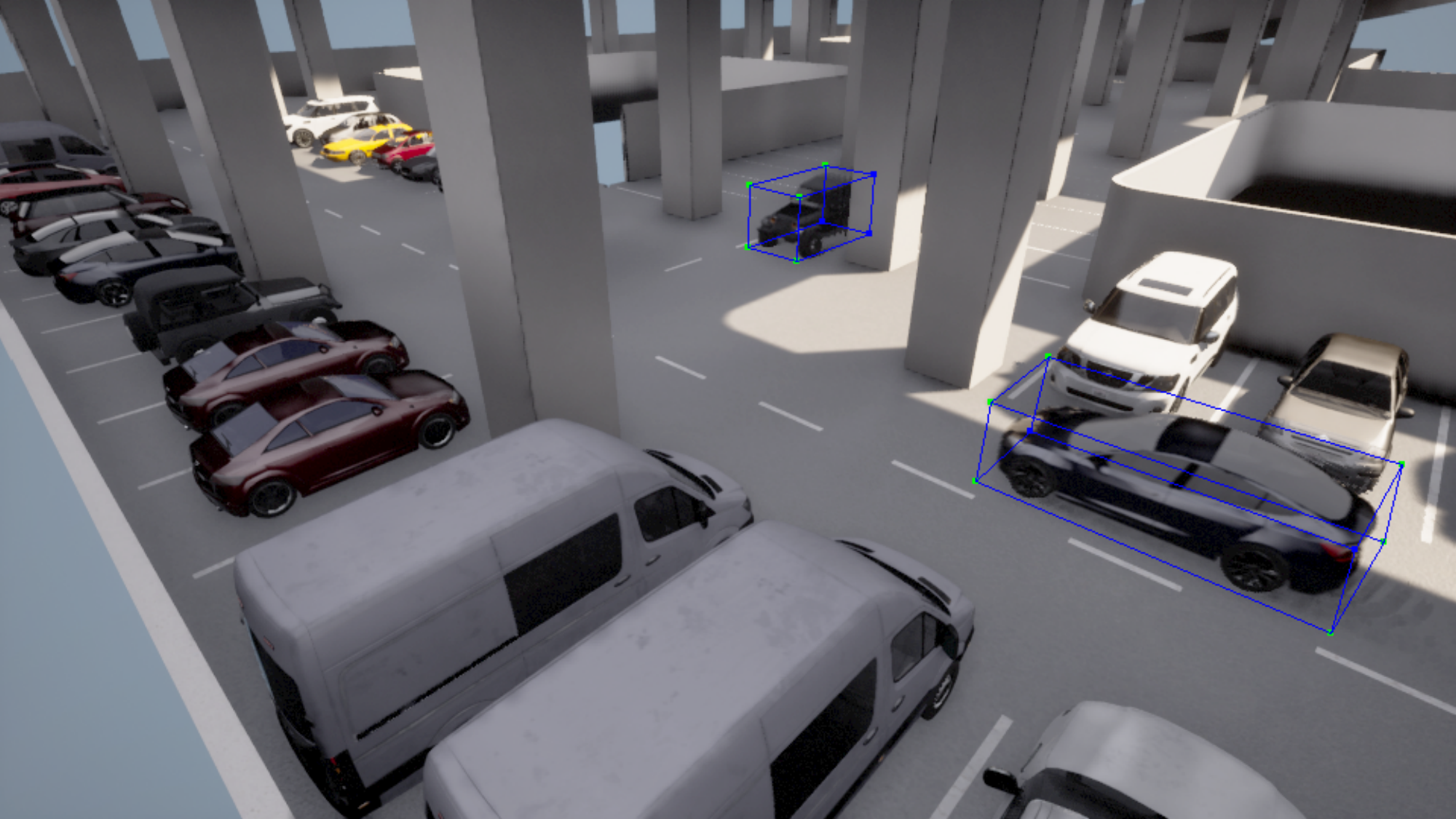}}
    \label{fig:s2_inf}
    \caption{Images from (a) the onboard camera and (b) the roadside camera in the T-junction scenario.}
    \label{fig:s2_images}
\end{figure}

\begin{table*}[htbp]
\centering
\caption{Number of Crashes in the T-junction Scenario}
\label{table:t-junc}
\renewcommand{\arraystretch}{1.2}
\begin{tabular}{|cc|cccccccc|}
\hline
\multicolumn{2}{|c|}{Sensor Configurations}                & \multicolumn{8}{c|}{Cruising Speed of Vehicles}   \\ \hline
\multicolumn{1}{|c|}{Veh. Sensors}              & Infra. Sensors & \multicolumn{1}{c|}{3.0m/s}                              & \multicolumn{1}{c|}{4.0m/s}                              & \multicolumn{1}{c|}{5.0m/s}                              & \multicolumn{1}{c|}{6.0m/s}                              & \multicolumn{1}{c|}{7.0m/s}                              & \multicolumn{1}{c|}{8.0m/s} & \multicolumn{1}{c|}{9.0m/s} & 10.0m/s \\ \hline
\multicolumn{1}{|c|}{}                             & None           & \multicolumn{1}{c|}{{\color[HTML]{FE0000} \textbf{0/20}}} & \multicolumn{1}{c|}{1/20}                                 & \multicolumn{1}{c|}{4/20}                                 & \multicolumn{1}{c|}{6/20}                                 & \multicolumn{1}{c|}{}                                  & \multicolumn{1}{c|}{}     & \multicolumn{1}{c|}{}     &       \\ \cline{2-10} 
\multicolumn{1}{|c|}{}                             & Cam.           & \multicolumn{1}{c|}{0/20}                                 & \multicolumn{1}{c|}{{\color[HTML]{FE0000} \textbf{0/20}}} & \multicolumn{1}{c|}{3/20}                                 & \multicolumn{1}{c|}{2/20}                                 & \multicolumn{1}{c|}{9/20}                                 & \multicolumn{1}{c|}{}     & \multicolumn{1}{c|}{}     &       \\ \cline{2-10} 
\multicolumn{1}{|c|}{\multirow{-3}{*}{Cam.}}       & Cam.+LiDAR     & \multicolumn{1}{c|}{}                                  & \multicolumn{1}{c|}{}                                  & \multicolumn{1}{c|}{0/20}                                 & \multicolumn{1}{c|}{{\color[HTML]{FE0000} \textbf{0/20}}} & \multicolumn{1}{c|}{3/20}                                 & \multicolumn{1}{c|}{5/20}    & \multicolumn{1}{c|}{7/20}    &       \\ \hline
\multicolumn{1}{|c|}{}                             & None           & \multicolumn{1}{c|}{0/20}                                 & \multicolumn{1}{c|}{{\color[HTML]{FE0000} \textbf{0/20}}} & \multicolumn{1}{c|}{1/20}                                 & \multicolumn{1}{c|}{5/20}                                 & \multicolumn{1}{c|}{5/20}                                 & \multicolumn{1}{c|}{}     & \multicolumn{1}{c|}{}     &       \\ \cline{2-10} 
\multicolumn{1}{|c|}{}                             & Cam.           & \multicolumn{1}{c|}{}                                  & \multicolumn{1}{c|}{0/20}                                 & \multicolumn{1}{c|}{{\color[HTML]{FE0000} \textbf{0/20}}} & \multicolumn{1}{c|}{3/20}                                 & \multicolumn{1}{c|}{7/20}                                 & \multicolumn{1}{c|}{3/20}    & \multicolumn{1}{c|}{}     &       \\ \cline{2-10} 
\multicolumn{1}{|c|}{\multirow{-3}{*}{Cam.+LiDAR}} & Cam.+LiDAR     & \multicolumn{1}{c|}{}                                  & \multicolumn{1}{c|}{}                                  & \multicolumn{1}{c|}{}                                  & \multicolumn{1}{c|}{0/20}                                 & \multicolumn{1}{c|}{{\color[HTML]{FE0000} \textbf{0/20}}} & \multicolumn{1}{c|}{4/20}    & \multicolumn{1}{c|}{2/20}    & 5/20     \\ \hline
\end{tabular}
\end{table*}

The second scenario happens at an internal T-junction of the garage, where another vehicle is turning left, but it is unaware of the ego AVP vehicle due to the occlusion of a pillar.
The objective of the ego AVP vehicle is to avoid collision by detecting the occluded vehicle and braking early enough.
The cruising speed of the turning vehicle is set to the same as the ego vehicle. 
The spawning positions of vehicles cause randomness in this scenario.

As shown in Fig. \ref{fig:s2_images}, severe occlusion is caused by the pillar near the T-junction, but it is well resolved with the aid of the infrastructure.
The results of the safety tests in the T-junction scenario are shown in Tab. \ref{table:t-junc}.
Unlike the pedestrian crossing scenario, the maximum safe cruising speed with and without onbaord LiDAR is 3m/s and 4m/s respectively, with a slight difference.
Besides, while the roadside camera improves the maximum safe speed by 1m/s, the safe speed is increased again by a large margin of 3m/s when the LiDAR is also included as the infrastructure.
The experiments necessitate the deployment of roadside sensors for infrastructure-assisted CP, which leads to a more reliable and efficient smart parking garage.

% \subsection{Effects of Communication Latency}

\section{Conclusions} \label{sect:conclusion}
In this work, we have proposed a BEV feature-based CP architecture in the framework of an infrastructure-assisted AVP system.
The size of transmitted features is effectively compressed to fit in the available data rate of the NR-V2X network. 
In two typical safety-critical scenarios, we have verified the effectiveness of infrastructure-assisted CP, which can increase the maximum safe cruising speed inside the parking garage by up to 3m/s, which equals 10.8km/h.
Future directions include studying the optimal deployment position and configuration of infrastructures in a parking garage and solving imperfect factors such as communication delays and localization errors.

\bibliographystyle{IEEEtran}

\end{document}